\documentclass[10pt,twocolumn,letterpaper]{article}

\usepackage{wacv}              %

\newcommand{\red}[1]{{\color{red}#1}}

\definecolor{wacvblue}{rgb}{0.21,0.49,0.74}
\usepackage[pagebackref,breaklinks,colorlinks,allcolors=wacvblue]{hyperref}
\usepackage{pifont}
\usepackage[inkscapelatex=false]{svg}
\usepackage[nolist]{acronym}
\usepackage[dvipsnames]{xcolor}
\usepackage{makecell}
\usepackage{graphics}
\usepackage{amssymb}
\usepackage{tablefootnote}
\usepackage{wasysym}
\usepackage[table]{xcolor}
\usepackage[accsupp]{axessibility}
\usepackage{multirow}

\newcommand{\green}[1]{\textcolor{Green}{#1}}

\definecolor{lightergray}{gray}{0.9}

\newcommand\blfootnote[1]{%
  \begingroup
  \renewcommand\thefootnote{}\footnote{#1}%
  \addtocounter{footnote}{-1}%
  \endgroup
}

\def\wacvPaperID{1443} %
\def\confName{WACV}
\def\confYear{2026}

\title{DenseBEV: Transforming BEV Grid Cells into 3D Objects}

\author{Marius Dähling$^{1,2,*}$
\and Sebastian Krebs$^{2,3}$
\and J. Marius Zöllner$^{1,4}$
}

\begin{acronym}
\acro{NMS}{Non-Maximum Suppression}
\acro{IoU}{Intersection-over-Union}
\acro{FoV}{Field of View}
\acro{mAP}{mean Average Precision}
\acro{NDS}{nuScenes detection score}
\acro{ATE}{Average Translation Error}
\acro{ASE}{Average Scale Error}
\acro{AOE}{Average Orientation Error}
\acro{AVE}{Average Velocity Error}
\acro{AAE}{Average Attribute Error}
\acro{LET}{Longitudinal Error Tolerant}
\acro{APL}{Average Precision Longitudinal}
\acro{APH}{Average Precision Heading}
\acro{LSS}{Lift, Splat, Shoot}
\acro{BEV}{Bird’s-Eye-View}
\end{acronym}

\begin{document}
\maketitle
\begin{abstract}
In current research, \ac{BEV}-based transformers are increasingly utilized for multi-camera 3D object detection. 
Traditional models often employ random queries as anchors, optimizing them successively.
Recent advancements complement or replace these random queries with detections from auxiliary networks. 
We propose a more intuitive and efficient approach by using \ac{BEV} feature cells directly as anchors.
This end-to-end approach leverages the dense grid of \ac{BEV} queries, considering each cell as a potential object for the final detection task.
As a result, we introduce a novel two-stage anchor generation method specifically designed for multi-camera 3D object detection. 
To address the scaling issues of attention with a large number of queries, we apply \ac{BEV}-based \acl{NMS}, allowing gradients to flow only through non-suppressed objects.
This ensures efficient training without the need for post-processing.
By using BEV features from encoders such as BEVFormer directly as object queries, temporal \ac{BEV} information is inherently embedded. 
Building on the temporal \ac{BEV} information already embedded in our object queries, we introduce a hybrid temporal modeling approach by integrating prior detections to further enhance detection performance.
Evaluating our method on the nuScenes dataset shows consistent and significant improvements in NDS and mAP over the baseline, even with sparser \ac{BEV} grids and therefore fewer initial anchors.
It is particularly effective for small objects, enhancing pedestrian detection with a $3.8\%$ mAP increase on nuScenes and an $8\%$ increase in LET-mAP on Waymo.
Applying our method, named DenseBEV, to the challenging Waymo Open dataset yields state-of-the-art performance, achieving a LET-mAP of $60.7\%$, surpassing the previous best by $5.4\%$.
Code is available at \url{https://github.com/mdaehl/DenseBEV}.
\end{abstract}
    
\section{Introduction}
\begin{figure}[t]
    \centering
    \includegraphics[width=\linewidth]{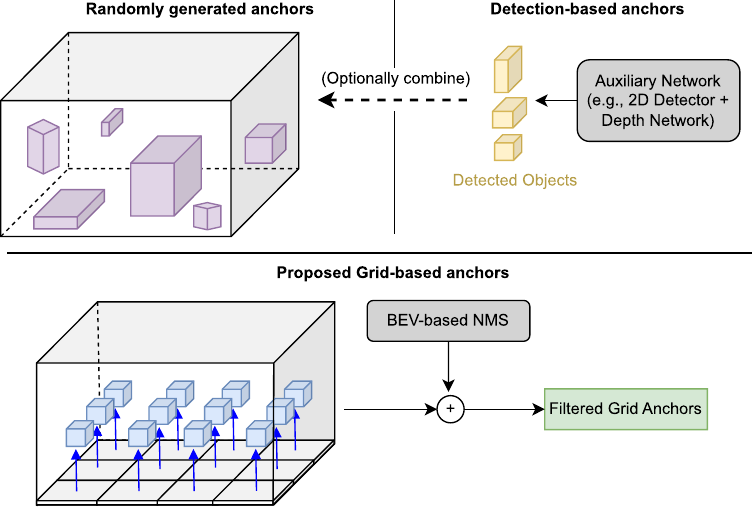}
    \caption{Comparison of anchor generation methods for BEV-based transformers in multi-camera 3D object detection. Previous methods (see upper half) use randomly generated anchors \cite{Zhiqi2022bev} or employ detection-based anchors that either replace \cite{jiang2024far3d} or complement \cite{yang2023bevformer} randomly generated anchors. Our method employs a BEV feature grid as anchors and refines them using NMS.}
    \label{fig:anchor}
\end{figure}

3D object detection is essential for applications such as autonomous driving and robotics.
\blfootnote{\hspace{-15pt}$^*$ Corresponding author: marius.daehling@mercedes-benz.com\\ 
$^1$Karlsruhe Institute of Technology (KIT)\\ 
$^2$Mercedes-Benz AG, Research and Development\\ 
$^3$Intelligent Vehicles Group at TU Delft\\
$^4$Research Center for Information Technology (FZI)}
Recently, many methods \cite{Zhiqi2022bev, wang2023exploring, liu2023sparsebev} have shifted from LiDAR to camera-based systems, driven by the lower cost and rich semantic information that cameras provide. 
This is particularly evident in autonomous driving, where multi-camera setups need to capture a comprehensive view of the surroundings.
Accurate detection of traffic participants and a comprehensive understanding of the environment are critical to enable safe autonomous driving. 
This includes not only robust detection of large objects like vehicles but also smaller objects such as pedestrians, bicycles, or barriers, which are equally important for safety. 
At the same time, improvements in detection performance must be balanced with computational efficiency, as real-world deployment remains a key constraint.

Effectively detecting a wide range of object classes, particularly smaller ones, requires careful prior selection.
In transformer-based detectors, these priors are represented by object queries, whose initialization strongly impacts final detection quality \cite{yang2023bevformer, ji2024enhancing}.
Early methods used randomly initialized queries that were either partially \cite{carion2020end} or fully learned \cite{zhu2020deformable}.
Recent methods replace or complement them with anchors from auxiliary networks \cite{yang2023bevformer, jiang2024far3d, ji2024enhancing}, often relying on smaller detectors to generate priors.
However, the quality of the final detection remains constrained by the capabilities of these auxiliary networks. 
If the initial detector fails to identify objects in difficult scenarios, the main model may struggle to recover them. 
Moreover, the use of additional detectors increases computational overhead, leading to a trade-off between anchor quality and efficiency.

In this paper, we introduce a novel method for initializing object queries in 3D object detection that maintains a low parameter count and adds minimal computational overhead.
 Unlike previous approaches, our method improves anchor generation over randomly initialized queries without relying on auxiliary networks (see Figure \ref{fig:anchor}).
We directly leverage the dense grid of rich environmental features from the \ac{BEV} encoder to generate initial object queries. 
This dense anchoring strategy enhances the detection of smaller objects that are often missed. 
However, dense anchors can introduce duplicates and increase computational costs. 
To address this, we incorporate \ac{NMS} into the training procedure, effectively filtering redundant queries.
Our object query formulation implicitly captures temporal context, as the \ac{BEV} encoder integrates past \ac{BEV} features. 
By incorporating previous detections, we merge these temporal \ac{BEV} features with temporal object information. 
This enriches the object queries with both spatial and temporal context.
We term this strategy hybrid temporal modeling, enabled by our novel object query formulation. We summarize our main contributions as follows:
\begin{enumerate}
    \item We present \textit{DenseBEV}, which, to the best of our knowledge, is the first end-to-end multi-view 3D object detection method to enable dense priors via \ac{NMS} while maintaining computational efficiency.
    \item We enable merging both temporal object-centering modeling and \ac{BEV} temporal modeling in the object queries, resulting in a hybrid temporal modeling scheme.
    \item Our method achieves state-of-the-art performance on the Waymo Open dataset \cite{sun2020scalability} and consistently improves on the base model across multiple model scales on the nuScenes dataset \cite{caesar2020nuscenes}.
\end{enumerate}

\section{Related Work}
We begin by reviewing multi-view camera-based 3D object detection methods in Section \ref{subsection:camera}, covering \ac{BEV}-based approaches, including transformers that operate on \ac{BEV} features as well as object-centric transformers. 
Section \ref{subsection:query} examines how object queries in transformer-based detectors are initialized and learned within these frameworks.

\subsection{Camera-based Multi-view 3D Object Detection}\label{subsection:camera}
Recent works \cite{wang2022mv, philion2020lift, yang2023bevformer, wang2023exploring} have explored multi-view 3D object detection using cameras. 
Some methods, like MV-FCOS3D++ \cite{wang2022mv} extend monocular approaches such as FCOS3D \cite{wang2021fcos3d}. 
However, the majority of recent methods focus on learning a unified spatial representation, such as a \ac{BEV}, that fuses information across multiple camera views.

Many approaches rely on \ac{BEV} features for 3D object detection, with various methods focusing on the generation of meaningful \ac{BEV} feature representations \cite{roddick2020predicting, philion2020lift, saha2022translating, yang2023bevformer, harley2023simple}. 
A pivotal work is \ac{LSS} \cite{philion2020lift}, which predicts a depth distribution per pixel and a corresponding context vector that are multiplied, resulting in a feature point cloud.
The point cloud is aggregated into \ac{BEV} space using PointPillars \cite{lang2019pointpillars}.
BEVDet \cite{huang2021bevdet} was one of the first to utilize \ac{LSS}, pairing it with a 3D LiDAR detection head.
BEVDet4D \cite{huang2022bevdet4d} added BEV temporal modeling with features from previous frames. 
BEVDepth \cite{li2023bevdepth} introduced a LiDAR-supervised depth estimation module to address poor depth prediction in \ac{LSS}. 
BEVNext \cite{li2024bevnext} improved depth prediction consistency using conditional random fields.
Recently, GeoBEV \cite{zhang2025geobev} advances \ac{LSS} by generating \ac{BEV} feature maps per image before merging them, thus reducing the sampling overhead.

Transformer-based models also emerged in this domain.
BEVFormer \cite{yang2023bevformer} defines a voxel space, projects each position into the images, samples features via deformable attention, and then collapses the voxel space to \ac{BEV}.
In addition, they leverage BEV temporal modeling.
SparseBEV \cite{liu2023sparsebev} avoids using a \ac{BEV} feature map by defining pillar queries in \ac{BEV} space that directly interact with the image feature maps.
The PETR series, including PETR \cite{liu2022petr}, PETRv2 \cite{liu2023petrv2}, and StreamPETR \cite{wang2023exploring}, introduces transformers that are object-based rather than \ac{BEV}-based. 
StreamPETR introduced object-centric temporal modeling, which was simultaneously presented by HoP \cite{zong2023temporal}.

Generally, object-based detection methods are more efficient but may lack semantic background information. 
Our work aims to enhance \ac{BEV}-based transformers by focusing on efficiency and incorporating concepts from object-based detection, including the object-centered temporal modeling.

\begin{figure*}[t]
    \centering
    \includegraphics[width=0.75\linewidth]{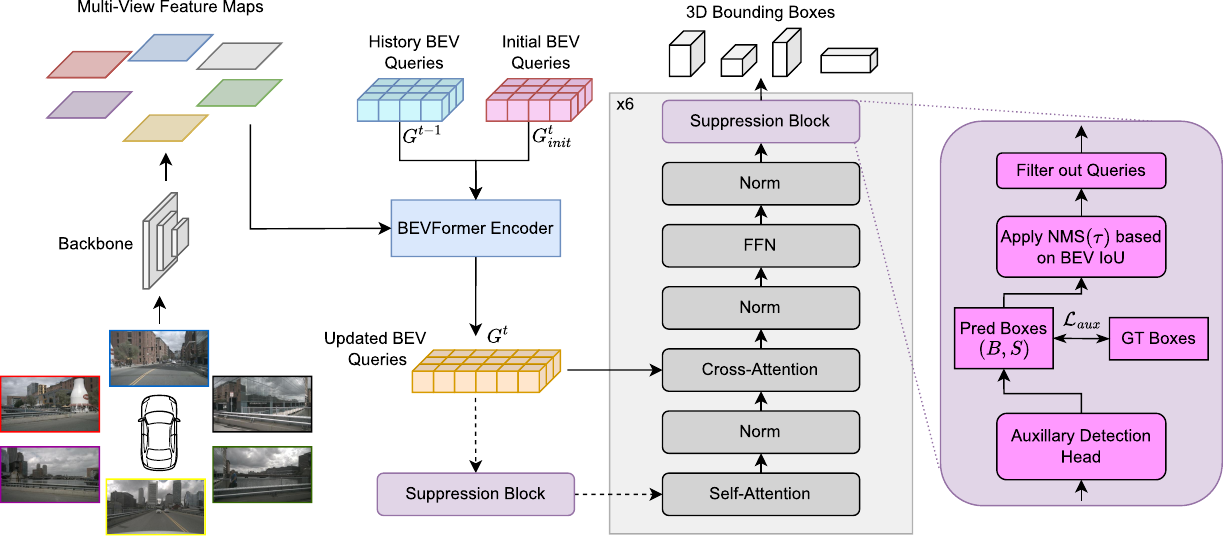}
    \caption{Overall architecture of our proposed method, \textit{DenseBEV}. Images are processed by a backbone network and encoded using the BEVFormer encoder \cite{Zhiqi2022bev}. Instead of introducing additional object queries, \textit{DenseBEV} utilizes the encoder output as object queries after passing them through the suppression block. To minimize duplicate predictions, a suppression block is placed after each decoder layer. The dashed line indicates that the gradient of the object queries is detached from the encoder.}
    \label{fig:architecture}
\end{figure*}

\subsection{Object Queries}\label{subsection:query}
DETR \cite{carion2020end} was the first transformer-based 2D object detector to introduce object queries.
The object queries are optimized in the decoder through refinement using encoder information.
These queries are divided into content and positional parts \cite{meng2021conditional}, with DETR \cite{carion2020end} using a frozen zero-vector as content part and learning only the positional part.

Deformable DETR \cite{zhu2020deformable} improves upon DETR by making content queries learnable embeddings and introducing iterative bounding box refinement.
The work also presents a two-stage method where object queries are initialized by decoding the encoder output and retraining the best boxes based on confidence score.
Subsequent works reformulated object queries as geometric anchors, with \cite{wang2022anchor} using anchor points $(x,y)$ and \cite{liu2022dabdetr} extending this to anchor boxes $(x,y,h,w)$.
DINO-DETR \cite{zhang2023dino} builds on DAB-DETR \cite{liu2022dabdetr} with a two-stage method, using the encoder output only for positional queries. The content queries are defined as learnable embeddings.
Although DDQ \cite{zhang2023dense} also uses a two-stage approach, they forward image features instead of bounding boxes as object queries to the decoder.
To limit the total number, all features are decoded and filtered using \ac{NMS}.

In 3D object detection, methods increasingly adopt two-stage approaches.
These rely on additional 2D \cite{jiang2024far3d, ji2024enhancing} or 3D detectors \cite{yang2023bevformer} to initialize the anchors. 
Far3D \cite{jiang2024far3d} uses a 2D detector and a depth network to lift 2D boxes to 3D. 
Similarly, QAF2D \cite{ji2024enhancing} creates 3D query anchors by combining detected 2D boxes with a predefined set of orientations and depths.
BEVFormer v2 \cite{yang2023bevformer} adds a perspective head to predict 3D proposals per image for the decoder.
In contrast, DAT \cite{zhang2023introducing} adds a depth network and rescale-based query selection, where queries are initialized based on downsampled and decoded \ac{BEV} features, followed by a top-k selection.

In summary, while two-stage approaches are used in both 2D and 3D detection, existing 3D methods mostly depend on auxiliary networks to produce initial anchors.
DAT avoids this by leveraging \ac{BEV} features as object queries, but its score-based filtering favors duplicate detections of easy objects while neglecting harder ones, and the reduced number of anchors limits performance in dense scenes, where each anchor can represent only a single object.
By contrast, our method removes the need for a separate network while maintaining dense priors, enabling strong performance with minimal overhead.

\section{Method}
Section \ref{section:two_stage} introduces our two-stage object query initialization mechanism for \ac{BEV}-based 3D object detection, leading to the \textit{DenseBEV} architecture shown in Figure \ref{fig:architecture}.
Section \ref{section:hybrid} then extends this framework with hybrid temporal modeling, resulting in the enhanced \textit{DenseBEV++} network.
While DenseBEV++ represents the complete model, we introduce DenseBEV as a standalone variant for scenarios where temporal information is unavailable or limited.

\subsection{Two-stage BEV Object Queries}\label{section:two_stage}
Inspired by the two-stage object query initialization of Deformable DETR \cite{zhu2020deformable}, we present two-stage \ac{BEV} object queries.
We propose to use features of the BEVFormer \cite{Zhiqi2022bev} encoder as initial object queries (anchors) in the decoder.
These features contain scene information, including details about the objects.
Each cell in the \ac{BEV} query grid acts as an anchor for a potential object that is ultimately detected.

Given the output of the BEVFormer \cite{Zhiqi2022bev} encoder, we initialize a grid of BEV queries $G^t \in \mathbb{R}^{n \times m \times C}$, where $n$ and $m$ define the grid resolution and $C$ the latent dimension.
Each cell is decoded using a simple auxiliary detection head, yielding $n \cdot m$ boxes.
Since the number of objects per scene is significantly lower than the number of cells in the grid, many predictions are redundant.
Moreover, with a regular BEV grid size of $200 \times 200$, this would yield 40,000 object queries, which is computationally impractical.
Therefore, it is crucial to reduce the number of queries before passing them to the decoder.
Following the approach of the 2D detector DDQ \cite{zhang2023dense}, we adapt the concept of suppressing duplicates using \ac{NMS} during training and extend it to 3D.
Performing \ac{NMS} in 3D is considerably more complex, as the boxes are rotated and thus not aligned with each other.
Additionally, the 3D Euclidean space is much larger than the 2D image space.
We mitigate this by applying \ac{NMS} in the \ac{BEV} plane, which both increases relative object density and simplifies computation.
This approach aligns with the assumption that objects in the context of autonomous driving are rarely stacked on top of each other.
Nonetheless, the \ac{IoU} threshold requires careful adjustment, since the objects are still more sparsely distributed compared to the 2D image space.

\begin{figure}
    \centering
    \includegraphics[width=0.9\linewidth]{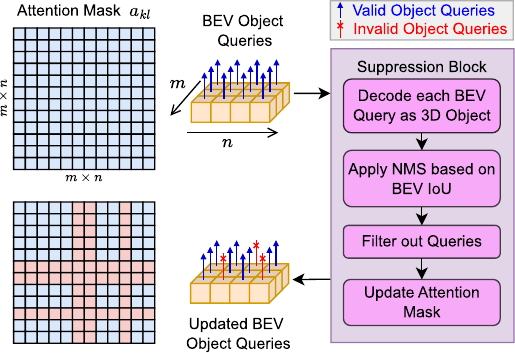}
    \caption{Visualization of attention mask. Initially, all object queries can attend to each other (blue squares), but after the decoding and \ac{NMS} process, suppressed ones are marked as invalid (red squares) in the attention mask.}
    \label{fig:mask}
\end{figure}

Formally, we start with the grid of \ac{BEV} queries $G^t$, which are decoded by auxiliary detection heads into a set of bounding boxes \mbox{$B=\{b_1, b_2, ..., b_{m \cdot n}\}$} with respective confidence scores \mbox{$S=\{s_1, s_2, ..., s_{m \cdot n}\}$}. 
All heads follow the BEVFormer \cite{Zhiqi2022bev} architecture, as it is simple and effective, with each head having its own weights.
The \ac{NMS} filtering to select relevant bounding boxes is defined as follows:
\begin{equation}
    B' = \text{BEV-NMS}(B, S, \tau),    
\end{equation}
where $\tau$ defines the \ac{IoU} threshold used for suppression.
Generally, the lower the \ac{IoU} threshold, the more objects will be suppressed.
We keep track of the suppressed objects using an attention mask that is progressively updated throughout the decoder layers.
While the attention mask is shared, each decoder layer uses its own auxiliary detection head.
The attention mask at position ($k,l$) is calculated as
\begin{equation}
    a_{kl} = 
    \begin{cases} 
        1, & \text{if } k \notin B' \text{ or } l \notin B' \\
        0, & \text{otherwise }.
    \end{cases}
\end{equation}
A value of 1 indicates that the attention is suppressed at that position, whereas a value of 0 indicates the opposite.
This is visualized in Figure \ref{fig:mask}. 
Following the initial suppression block before the decoder, a top-k selection based on confidence scores is performed to maintain a consistent input size.
The top-k selection is not required for the suppression blocks inside the decoder.

We avoid using class-aware \ac{NMS} or Scale \ac{NMS}\cite{huang2021bevdet}, which scales objects based on the class before intersection calculation, due to the necessity of high-quality class predictions.
The likelihood of faulty class detections, particularly in earlier layers, is significant and could be detrimental (see supplementary \ref{subsection:class_aware}). 
Consequently, we choose not to use class-based dimension anchors, as outlined in QAF2D\cite{ji2024enhancing}.

We do not add positional encoding to the object queries before feeding them into the decoder, since it is already included in the encoder.
All objects are initially placed at the center height of the pre-defined detection space. 

\subsection{Hybrid Temporal Modeling}\label{section:hybrid}
Our novel object query formulation (Section \ref{section:two_stage}) enables the integration of temporal \ac{BEV} information directly into the object queries. 
We argue that extending this temporal context with historical \ac{BEV} features further benefits the 3D object detection task. 
To this end, we enhance our two-stage \ac{BEV} object queries by incorporating previously detected objects as additional inputs.

Inspired by Stream-PETR \cite{wang2023exploring}, we maintain a memory queue that stores past detections across multiple time steps, updated in a first-in, first-out (FIFO) manner. 
By combining the historical \ac{BEV} features in our object queries with the temporal information from this memory queue, we introduce a hybrid temporal modeling approach that leverages the complementary strengths of both temporal sources.

\begin{figure}
    \centering
    \includegraphics[width=0.8\linewidth]{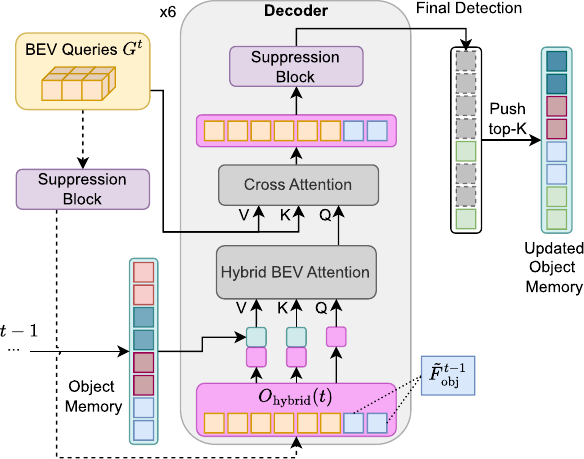}
    \caption{Hybrid Temporal Modeling. Detections from the previous timestamp, shown in different colors in the object memory, are concatenated with current \ac{BEV} queries to form extended object queries. A memory queue containing detections from multiple earlier timestamps is also incorporated into the keys and values of subsequent attention layers. For clarity, motion layer normalizations from StreamPETR \cite{wang2023exploring} are omitted.}
    \label{fig:hybrid}
\end{figure}

To ensure consistency, we treat the prior detections in the memory queue in the suppression block in the same manner as the \ac{BEV} object queries.
Since these detections offer strong spatial priors, adjusting them is often easier and more reliable than predicting objects from scratch. 
All queries, both from the \ac{BEV} grid and the previously detected objects, are included in the loss computation. 
Special care is taken to correctly integrate the additional queries into the attention mechanism, particularly within the suppression block.
The overall process is illustrated in detail in Figure \ref{fig:hybrid}.

To provide a comprehensive understanding of our approach, we present the details subsequently. 
We adopt notation similar to StreamPETR \cite{wang2023exploring}, with formulas simplified to account for only one previous timestamp. 
Intermediate layers are omitted to focus on the temporal formulation. 
Equations unroll recursively for more than one timestamp.
We begin with the \ac{BEV} temporal modeling, which is utilized in our method for object queries and cross-attention in the decoder. 
The \ac{BEV} features from the current step, incorporating temporal information, are defined as:
\begin{equation}
   {G}^{t} =  \varphi(G^{t-1}, G^{t}_{\text{init}}),
\end{equation}
where $\varphi$ represents temporal self-attention, $G^{t-1}$ denotes previous \ac{BEV} queries, and $G^{t}_{\text{init}}$ refers to initial \ac{BEV} queries from the current timestep.
In the object-centric modeling, the features containing information about objects, including temporal data, are expressed as:
\begin{equation}
    \tilde{F}^{t-1}_{\text{obj}} = \mu(F^{t-1}_{\text{obj}}, M),
\end{equation}
where $F^{t-1}_{obj}$ are features of previously detected objects and $\mu$ is an implicit function to encode the motion attributes $M$.
By concatenating both feature sets, we define the object queries for the current timestamp as:
\begin{equation}
    O_{\text{hybrid}}(t) = {G}^{t} \mathbin{\|} \tilde{F}^{t-1}_{\text{obj}},
\end{equation}
which concatenates both types of queries.

Finally, we integrate the \textit{look forward twice} scheme from DINO-DETR \cite{zhang2023dino} to enhance information exchange between the two different types of object queries across more than a single decoder layer.

\subsection{Training}
Training is straightforward, with the only change compared to the baseline being the loss calculation.
The overall loss, considering the auxiliary detection heads, is defined as:
\begin{equation}
    \mathcal{L}_{\text{total}} = \mathcal{L}_{\text{base}} + \sum_i \lambda_i \mathcal{L}_{\text{aux}_i}
\end{equation}
where $\mathcal{L}_{\text{aux}_i}$ represents the output of the $i$th auxiliary detection head.
This includes the detection head prior to the decoder. Each auxiliary loss is weighted by $\lambda_i$.
To account for \ac{NMS} during training, the attention mask controls which boxes contribute to the loss calculation.

\section{Experiments}
\begin{table*}[t]
    \centering
    \resizebox{0.85\linewidth}{!}{
    \begin{tabular}{lc|cc|ccccc}
        \toprule
        \textbf{Method} & \textbf{BEV Size}  & \textbf{NDS} $\uparrow$ & \textbf{mAP} $\uparrow$ & \textbf{mATE} $\downarrow$ & \textbf{mASE} $\downarrow$ & \textbf{mAOE} $\downarrow$ & \textbf{mAVE} $\downarrow$ & \textbf{mAAE} $\downarrow$ \\
        \midrule
        BEVFormer-tiny* & $50 \times 50$  & 38.7  & 27.3 & 0.869 & \textit{0.285}  & 0.625 & 0.509  & 0.209 \\
        \rowcolor{lightergray}
        DenseBEV-tiny &  $50 \times 50$  & \textit{40.4}  & \textit{27.8}  & \textit{0.839} & \textbf{0.284}  & \textbf{0.549} & \textbf{0.474}  & \textit{0.208}\\
        \rowcolor{lightergray}
        DenseBEV++-tiny &  $50 \times 50$  & \textbf{41.3}  & \textbf{30.0}  & \textbf{0.824} & 0.285  & \textit{0.583} & \textit{0.478}  & \textbf{0.204} \\
        \midrule
        BEVFormer-small* & $150 \times 150$  & 50.1   & 39.5  & 0.694 & 0.275  & 0.386 & 0.409  & \textit{0.200} \\
        \rowcolor{lightergray}
        DenseBEV-small & 1$50 \times 150$  & \textit{51.0}  &\textit{40.5}  &\textit{0.670} & \textit{0.272}  & \textit{0.371} & 0.414  & 0.200 \\
        \rowcolor{lightergray}
        DenseBEV++-small &  $150 \times 150$  & \textbf{52.8}  & \textbf{42.8}  & \textbf{0.664} & \textbf{0.270}  & \textbf{0.338} & \textbf{0.395}  & \textbf{0.192} \\
        \midrule
        PolarFormer \cite{jiang2023polarformer} & $256 \times 64$ $\dagger$ & 52.8 & 43.2 & 0.648 & \textit{0.270} & 0.348 & 0.409 & 0.201 \\
        DAT-BEVFormer \cite{zhang2023introducing} & $200 \times 200$ & \textit{54.5} & 43.3 & \textit{0.623} & 0.271 & 0.351 & 0.309 & \textbf{0.188} \\
        BEVFormer-base  & $200 \times 200$  & 51.7  & 41.6  & 0.672 & 0.274  & 0.369 & 0.397  & 0.198 \\
        \rowcolor{lightergray}
        DenseBEV-base & $200 \times 200$ & 53.5  & \textit{43.3}  & 0.638 & 0.271  & \textit{0.344} & \textit{0.375}  & \textit{0.189} \\
        \rowcolor{lightergray}
        DenseBEV++-base & $200 \times 200$ & \textbf{54.9}  & \textbf{44.9}  & \textbf{0.615} & \textbf{0.264}  & \textbf{0.330} & \textbf{0.360}  & 0.189 \\
        \bottomrule
    \end{tabular}}
    \caption{Comparison on nuScenes val set. $\dagger$ highlights, that the BEV grid is in polar coordinates, hence cells are not quadratic. * refers to retrained models without the rotation bug (see Section \ref{subsection:results}). Bold indicates the best results, and italics denote the second best. This convention applies to all other tables as well.}
    \label{table:nusc}
\end{table*}
\begin{table}[t]
    \centering
    \resizebox{\columnwidth}{!}{
    \begin{tabular}{lc|ccc}
        \toprule
        \textbf{Method} & \textbf{Backbone} & \textbf{mAPL} $\uparrow$ & \textbf{mAP} $\uparrow$ & \textbf{mAPH} $\uparrow$ \\
        \midrule
        PETRv2$\dagger$ \cite{liu2023petrv2} & R101 & 36.6 & 51.9 & 47.9 \\
        StreamPETR$\dagger$ \cite{wang2023exploring} & R101  & 39.9  & 55.3  & 51.7 \\
        MV-FCOS3D++ \cite{wang2022mv} & R101-DCN  & 38.5  & 53.2  & 50.0 \\
        BEVFormer \cite{Zhiqi2022bev}  & R101-DCN  & 38.2  & 55.1  & 51.1 \\
        \midrule
        \rowcolor{lightergray}
        DenseBEV & R101-DCN  & 41.1  & 58.3  & 54.6 \\
        \rowcolor{lightergray}
        DenseBEV++ (w/o velocity) & R101-DCN  & \textbf{43.2}  & \textbf{60.7}  & \textbf{56.9} \\
        \rowcolor{lightergray}
        DenseBEV++ & R101-DCN  & 42.4  & 60.2 & 56.4  \\
        \bottomrule
    \end{tabular}
    }
    \caption{Comparison on Waymo val set. $\dagger$ indicates results are taken from \cite{wang2023exploring}.}
    \label{table:waymo}
\end{table}

\subsection{Experimental Setup}
\textbf{Datasets:}
We evaluate our method on the widely used nuScenes dataset \cite{caesar2020nuscenes} and the Waymo Open dataset \cite{sun2020scalability}.

The nuScenes dataset comprises 1000 scenes, each lasting 20 seconds, and is annotated at 2 Hz. 
The scenes are divided into 700 for training, 150 for validation, and 150 for testing.
The sensor setup includes 6 cameras providing a full 360° horizontal \ac{FoV}.
The object detection task involves 10 classes. %

In comparison, the Waymo Open dataset consists of 1150 scenes, split into 798 for training, 202 for validation, and 150 for testing.
Each scene spans 20 seconds, with annotations provided at a frequency of 10 Hz.
The images are captured by 5 cameras, each with a horizontal FOV of 50.4°. 
These cameras are slightly overlapping, leaving the rear section of the vehicle uncovered.
The object detection task includes 3 classes: vehicle, pedestrian, and cyclist. 

\textbf{Metrics:}
The primary evaluation metrics of the nuScenes dataset are the \ac{NDS} and the well-known \ac{mAP}.
Additionally, \ac{ATE}, \ac{ASE}, \ac{AOE}, \ac{AVE} and \ac{AAE} are calculated.
Ultimately, the \ac{NDS} combines object detection performance through the \ac{mAP} and the other metrics that account for location, size, orientation, velocity, and attribute quality.

The Waymo Open dataset evaluates camera-based object detectors using custom \ac{LET} metrics \cite{hung2024let}. 
These metrics are similar to the \ac{mAP} but allow for a longitudinal tolerance during the matching process to address depth prediction challenges in camera-based approaches.
The final evaluation includes the \ac{mAP} as well as \ac{APL} and \ac{APH}, which weigh the results based on longitudinal and heading errors, respectively.

\textbf{Implementation:}
Our implementation is built on MMDetection3D \cite{mmdet3d2020}, extending the original BEVFormer \cite{Zhiqi2022bev} codebase. 
The object memory functionality relies on StreamPETR \cite{wang2023exploring}. 
The \ac{NMS} computation for rotated boxes is non-trivial. 
Therefore, we employ the CUDA implementation from OpenPCDet \cite{openpcdet2020}, and improve runtime by adding heuristics.

\textbf{Training:}
We maintain 900 object queries and propagate 300 historical object queries per timestamp when using the hybrid temporal modeling functionality.
The \ac{IoU} threshold is set to $\tau=0.1$ for the base and small model. The tiny model uses $\tau=0.2$ to account for the sparser BEV grid.
The auxiliary losses are weighted equally to the final detection loss.
All other configurations, including the backbone choice, adhere to BEVFormer defaults.
Following common practice, we train on Waymo using every 5th image.
All experiments were conducted on 8 NVIDIA A100 GPUs.

\subsection{Results}\label{subsection:results}
\begin{table*}[t]
    \centering
    \scriptsize
    \resizebox{\linewidth}{!}{
        \begin{tabular}{lcccccccccccc}
        \toprule
        \vspace{-6pt}\\[-6pt]
        Method & \rotatebox{40}{\makecell{Traf. Cone \\ $\varnothing=0.19\,\mathrm{m}^3$}} &
        \rotatebox{40}{\makecell{Pedestrian \\ $\varnothing=0.87\,\mathrm{m}^3$}} &
        \rotatebox{40}{\makecell{Bicycle \\ $\varnothing=1.32\,\mathrm{m}^3$}} &
        \rotatebox{40}{\makecell{Barrier \\ $\varnothing=1.32\,\mathrm{m}^3$}} &
        \rotatebox{40}{\makecell{Motorcycle \\ $\varnothing=2.44\,\mathrm{m}^3$}} &
        \rotatebox{40}{\makecell{Car \\ $\varnothing=15.59\,\mathrm{m}^3$}} &
        \rotatebox{40}{\makecell{Truck \\ $\varnothing=62.45\,\mathrm{m}^3$}} &
        \rotatebox{40}{\makecell{Constr. Veh. \\ $\varnothing=79.57\,\mathrm{m}^3$}} &
        \rotatebox{40}{\makecell{Bus \\ $\varnothing=105.08\,\mathrm{m}^3$}} &
        \rotatebox{40}{\makecell{Trailer \\ $\varnothing=143.83\,\mathrm{m}^3$}} \\
        \midrule
        BEVFormer & 58.4 & 49.4 & 39.8 & 52.5 & 42.9 & \textit{61.8} & \textit{37.0} & 12.7 & \textit{44.4} & \textbf{17.2} \\
        \rowcolor{lightergray}
        DenseBEV & \textit{63.5} (\green{+5.1\%}) & \textit{51.7} (\green{+2.3\%}) & \textit{41.2} (\green{+1.4\%}) & \textit{59.2} (\green{+6.7\%}) & \textit{44.0} (\green{+1.1\%}) & 61.3 (\red{-0.5\%}) & 36.0 (\red{-1.0\%}) & \textbf{13.0} (\green{+0.3\%}) & 43.8 (\red{-0.6\%}) & 17.1 (\red{-0.1\%}) \\
        \rowcolor{lightergray}
        DenseBEV++ & \textbf{63.5} (\green{+5.1\%}) & \textbf{53.2} (\green{+3.8\%}) & \textbf{48.2} (\green{+8.4\%}) & \textbf{60.6} (\green{+8.1\%}) & \textbf{45.6} (\green{+2.7\%}) & \textbf{63.1} (\green{+1.3\%}) & \textbf{38.6} (\green{+1.6\%}) & \textit{12.9} (\green{+0.2\%}) & \textbf{46.2} (\green{+1.8\%}) & \textit{17.1} (\red{-0.1\%}) \\
        \bottomrule
        \end{tabular}
    }
    \caption{Class specific \ac{mAP} on nuScenes val. Columns are ordered by average object size from left (small) to right (large).}
    \label{tab:nuscenes_classes}
\end{table*}
\begin{table}[t]
    \centering
    \resizebox{\columnwidth}{!}{
        \begin{tabular}{lcccccc}
            \toprule
            & \multicolumn{3}{c}{Vehicle} & \multicolumn{3}{c}{Pedestrian} \\
            \cmidrule(lr){2-4} \cmidrule(lr){5-7}
            Method     & mAPL     & mAP    & mAPH   & mAPL      & mAP     & mAPH    \\
            \midrule 
            BEVFormer  & 52.5 & 71.1 & 69.6 & 34.3 & 51.3 & 43.9 \\
            \rowcolor{lightergray}
            DenseBEV & \textit{55.0} & \textit{74.3} & \textit{73.0} & \textit{39.4} & \textit{57.6} & \textit{50.4} \\
            \rowcolor{lightergray}
            DenseBEV++ &  \textbf{57.6} & \textbf{76.6} & \textbf{75.3} & \textbf{40.8} & \textbf{59.3} & \textbf{52.1} \\
            \bottomrule
        \end{tabular}
    }
    \caption{Class-specific improvements on Waymo. In Waymo, vehicles comprise about $68\%$ and pedestrians around $31\%$. Cyclists are excluded as they represent less than $1\%$.}
    \label{table:waymo_classes}
\end{table}

\textbf{nuScenes Dataset:}
The nuScenes experiments aim to demonstrate the substantial impact of our methods, highlighting their improvements over the baseline.

Retraining BEVFormer tiny and small locally produced better results than those reported in their paper \cite{Zhiqi2022bev}, using the same settings and hyperparameters.
Additionally, a bug\footnote{https://github.com/fundamentalvision/BEVFormer/issues/96} in the original BEVFormer code was identified that affects all models besides the base version.
The rotation center used to align the \ac{BEV} features between timesteps is fixed at $(100, 100)$, which leads to incorrect feature rotation for any resolution other than $200 \times 200$.
This bug impacts our model more severely than the baseline, as it directly affects our object queries.
To ensure a fair comparison, we present the scores of our retrained and bug-free models in the Table~\ref{table:nusc}.
We emphasize that all results for \textit{DenseBEV(++)} were obtained after fixing the bug, since the issue has a greater effect on our architecture.

We report the results on the NuScenes val set in Table~\ref{table:nusc}, including the baseline, our models, and related methods.
HoP \cite{zong2023temporal} is excluded because it focuses on the encoder, making it orthogonal to our decoder-focused approach.
As shown, our models enhance both the \ac{NDS} and \ac{mAP} across all model scales.
\textit{DenseBEV}-tiny increases the NDS by $1.7\%$, while the \ac{mAP} sees a small improvement of $0.5\%$. 
This indicates that primarily the quality of the detected objects improves, rather than the quantity of correctly predicted objects. 
With the low \ac{BEV} resolution, the number of anchors is only slightly higher than the number of object queries in the decoder. 
Incorporating hybrid temporal modeling results in balanced improvements, with a total increase of $2.6\%$ in \ac{NDS} and $2.7\%$ in \ac{mAP}. 
\textit{DenseBEV}-small and \textit{DenseBEV++}-small demonstrate balanced performance gains, with \textit{DenseBEV++} achieving a $2.7\%$ increase in \ac{NDS} and a $3.3\%$ increase in \ac{mAP}.
\textit{DenseBEV}-base demonstrates the largest improvement due to the superior \ac{BEV} features derived from higher grid resolution and additional encoder layers. 
These improved \ac{BEV} features directly translate to better anchors in our models. 
Ultimately, \textit{DenseBEV++}-base boosts the \ac{NDS} and \ac{mAP} by $3.2\%$ and $3.3\%$, respectively.
In addition to the improvements in the primary metrics, clear enhancements are particularly evident in mATE, mAOE, and mAVE.
\textit{DenseBEV++}-base improves mATE by $5.7\%$ compared to BEVFormer-base, attributed to the dense grid of anchor queries facilitating more precise localization.
In terms of mASE and mAAE, the results are generally comparable to the baseline, with sometimes minor improvements.
Beyond comparisons with BEVFormer, our model also outperforms related methods.
\textit{DenseBEV++}-base surpasses DAT on all metrics except mAAE (marginal difference) and does so without relying on an auxiliary network.
Against PolarFormer, which employs a polar grid, our approach further improves NDS by 2.1\% and mAP by 1.7\%.

\textbf{Waymo Open Dataset:}
The Waymo dataset experiments focus on comparing our models against other approaches, emphasizing our competitive performance.

Our first model, \textit{DenseBEV}, outperforms the baseline, with an increase of nearly $3\%$ across all metrics.
Specifically, the mAPL improves $2.9\%$, mAP by $3.2\%$, and mAPH by $3.5\%$.
As highlighted in Table \ref{table:waymo}, \textit{DenseBEV} already surpasses the previous state-of-the-art, StreamPETR \cite{wang2023exploring}. 
Notably, this is reflected in an improvement of $3.0\%$ in mAP and $2.9\%$ in mAPH.

Implementing hybrid temporal modeling on the Waymo dataset requires adjustments due to the absence of object velocity in the detection task.
Although velocity data is available, it is not originally utilized since the metrics do not evaluate it.
Therefore, we tested DenseBEV++ both with and without incorporating velocity into the detection task. 
The results are quite similar, with the version excluding velocity performing moderately better.
While incorporating velocity might enhance performance for other metrics that consider it, it does not provide any benefit for the standard metrics.
Nevertheless, the hybrid temporal modeling improves overall performance.
Compared to \textit{DenseBEV}, \textit{DenseBEV++} achieves improvements ranging from $2.1\%$ in mAPL to $2.4\%$ in mAP. 
Finally, \textit{DenseBEV++} surpasses the state-of-the-art by achieving improvements of $3.3\%$ in mAPL, $5.4\%$ in mAP, and $5.2\%$ in mAPH.

\begin{figure}[t]
    \centering
    \includegraphics[width=\linewidth]{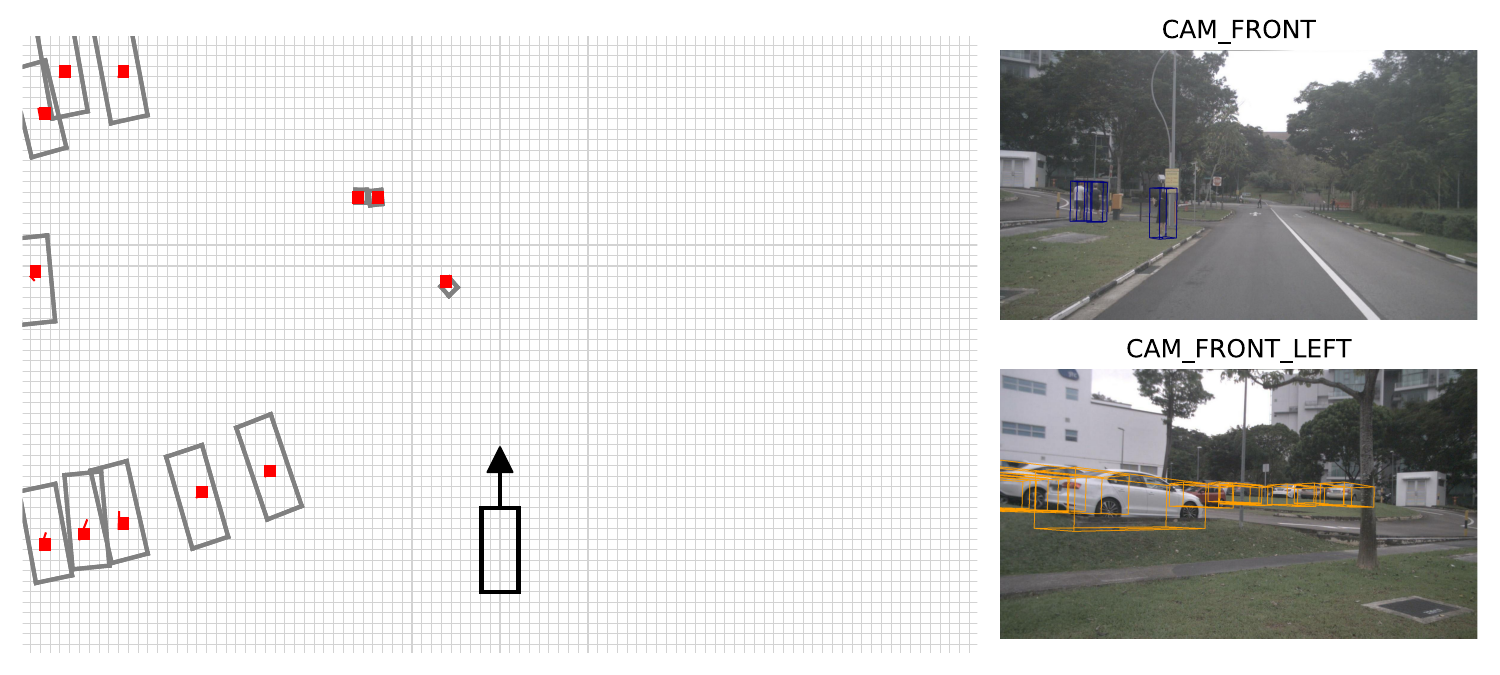}
    \caption{On the left, a BEV grid section near the ego vehicle (black box with forward arrow) is shown. The red square highlights the anchor grid position used to detect the corresponding bounding box, linked by a red line. The right side shows the related camera images for context.}
    \label{fig:grid}
\end{figure}

\textbf{Class specific improvements}:
To better understand the effect of dense anchors, we analyze performance across different object classes on the nuScenes and Waymo datasets. 
A central strength of our approach is its ability to improve detection of smaller objects, which are often the most challenging in autonomous driving scenarios.

As shown in Table~\ref{tab:nuscenes_classes}, \textit{DenseBEV} substantially improves small static objects such as traffic cones and barriers.
For larger classes, results are mostly comparable to the baseline, with minor decreases in some cases (see supplementary material \ref{subsection:large_obj} for details).
Importantly, \textit{DenseBEV++} further boosts performance on small objects, particularly dynamic ones like pedestrians and bicycles. 
Additionally, the detection performance on larger objects slightly improves compared to the baseline.

On the Waymo dataset (Table~\ref{table:waymo_classes}), both models consistently outperform the baseline, showing improvements for both object classes.
Consistent with our findings on the nuScenes dataset, we observe larger gains for the physically smaller pedestrian class compared to the larger vehicle class.
Nevertheless, \textit{DenseBEV++} improves all metrics for the vehicle class by more than $5\%$.
For pedestrians, all metrics improve by over $6\%$, with particularly notable increases of approximately $8\%$ in mAPL and mAP.

\textbf{Visualization:}
In Figure \ref{fig:grid}, we visualize the detections of DenseBEV++ on a section of a scene. 
It shows the detected boxes and illustrates how our method uses the grid cells within these boxes as anchors, aligning with our goal of selecting grid cells near object centers as anchors.
Qualitative results are provided in the supplementary material \ref{subsection:qualitative}.

\begin{table*}[ht]
    \centering
    \begin{subtable}[t]{0.2\linewidth}
        \centering
        \resizebox{\linewidth}{!}{
        \begin{tabular}[t]{l|cc}
        \toprule
        $\boldsymbol{\tau}$ & \textbf{NDS} $\uparrow$ & \textbf{mAP} $\uparrow$ \\
        \midrule
        0.1 & \textbf{53.5}  & \textbf{43.3} \\
        0.2 & \textit{53.5}  & \textit{43.3} \\
        0.3  & 52.6  & 42.3 \\
        0.5 & 52.8  & 42.7 \\
        0.8  & 51.7  & 41.6 \\
        \bottomrule
        \end{tabular}%
    }
    \caption{\ac{IoU} thresholds $\tau$.}
    \label{table:iou}
    \end{subtable}
    \hfill
    \begin{subtable}[t]{0.25\linewidth}
        \centering
        \resizebox{\linewidth}{!}{%
        \begin{tabular}[t]{c|cc}
        \toprule
        \textbf{BEV Size} & \textbf{NDS} $\uparrow$ & \textbf{mAP} $\uparrow$ \\
        \midrule
        $100 \times 100$ & 52.7  & 42.2 \\
        $150 \times 150$ & 52.8  & 42.6 \\
        $200 \times 200$ & \textit{53.5}  & \textit{43.3} \\
        $300 \times 300$ & \textbf{53.6}  & \textbf{43.8} \\
        \bottomrule
        \end{tabular}
        }
        \caption{\ac{BEV} grid resolutions.}
        \label{table:bev_res}
    \end{subtable}
    \hfill
    \begin{subtable}[t]{0.41\linewidth}
        \centering
        \resizebox{\linewidth}{!}{%
        \begin{tabular}[t]{cc|cc}
        \toprule
        \textbf{Memory Loss} & \textbf{Attention Mask} & \textbf{NDS} $\uparrow$ & \textbf{mAP} $\uparrow$ \\
        \midrule
        \ding{55} & \ding{55} & 52.9 & 40.5 \\
        \ding{55} & \ding{51} & \textit{54.7} & \textit{44.6} \\
        \ding{51} & \ding{55} & 54.3 & 44.1 \\
        \ding{51} & \ding{51} & \textbf{54.9} & \textbf{44.9} \\
        \bottomrule
    \end{tabular}
    }
    \caption{Different approaches to integrate object-centric memory.}
    \label{table:mask}
    \end{subtable}
    \caption{Performance Ablations on NuScenes val. We evaluated different IoU suppression thresholds \subref{table:iou}, BEV resolutions \subref{table:bev_res}, and methods to integrate object-centric memory \subref{table:mask}. The attention mask indicates whether queries are part of the BEV-NMS or not.}
    \label{tab:ablation}
\end{table*}

\subsection{Ablations}
All ablation experiments are conducted on the nuScenes dataset using the base size model, an \ac{IoU} threshold of \mbox{$\tau=0.1$} and without hybrid memory unless otherwise specified.
We start by comparing various \ac{IoU} thresholds to examine the influence of this hyperparameter. 
Subsequently, we assess the effect of different \ac{BEV} grid resolutions.
Furthermore, we explore various methods for integrating previously detected objects into the \ac{BEV}-based queries for the hybrid temporal modeling.
Finally, the computational overhead of our method is evaluated.

\textbf{IoU threshold:}
In Table \ref{table:iou} we compare various \ac{IoU} thresholds for the \ac{NMS}.
Generally, lower threshold values tend to result in better scores.
The importance of proper threshold selection is evident from the $1.8\%$ \ac{NDS} and $1.7\%$ \ac{mAP} difference between the settings $\tau=0.1$ and $\tau=0.8$.
However, performance appears to stabilize, as indicated by similar scores for $\tau=0.1$ and $\tau=0.2$.
These thresholds are significantly lower than those typically used in 2D object detection, due to the sparser nature of the \ac{BEV} space.
The setting $\tau = 0.1$ yields the best results, although the rounded results are identical to those of $\tau = 0.2$.
Ultimately, we choose $\tau=0.1$ as our final setting, prioritizing more aggressive suppression to minimize false positives.

\textbf{\ac{BEV} resolution:}
The original setting uses a \ac{BEV} resolution of $200 \times 200$. 
We note that the model with resolution $150 \times 150$ differs from the small model, as it employs six instead of three encoder layers.
As indicated by the numbers in Table \ref{table:bev_res}, increasing the resolution consistently improves model performance.
We also observe that decreasing the resolution affects the \ac{mAP} more than the \ac{NDS}.
Hence, the quality of the detected objects is less affected, but generally, finding correct boxes is.
Scaling the resolution beyond a certain point greatly increases computational cost, as the operations scale quadratically.
Although the densest grid of $300 \times 300$ yields the best results, it requires substantially more resources. 
The additional cost, with over $50\%$ longer training and nearly double the inference time, is not justified by the relatively minor improvement in performance.

\textbf{Object-centric memory:}
We investigate the optimal method for integrating object queries from previous detections. 
The different options include whether these queries participate in the final loss and whether they can suppress \ac{BEV} object queries by being part of the attention mask. As the data in Table \ref{table:mask} indicates, it is crucial to utilize at least one of these options. 
Therefore, active interaction between both types of queries is essential. 
Treating both types of object queries the same yields the best results.

\textbf{Computation efficiency:}
Table \ref{table:runtime} evaluates the efficiency of our method.
Replacing the learnable queries with our \ac{BEV} object queries surprisingly reduces the number of parameters compared to the baseline, although the reduction of $0.04\%$ is negligible. 
The runtime overhead of $18.5\%$ is primarily attributed to the \ac{NMS} calculation in the first stage prior to the decoder, as all cell objects are compared to each other.
Utilizing hybrid temporal modeling increases the parameters by only $0.55\%$ compared to the baseline, resulting in a minimal FPS drop of $0.4\%$ compared to using \ac{BEV} object queries alone.
While these findings confirm that the computational cost is moderate, we additionally investigate runtime reduction strategies for scenarios with limited compute in the supplementary (Section~\ref{subsection:ext_efficiency}).

\begin{table}[t]
    \centering
    \resizebox{\columnwidth}{!}{
    \begin{tabular}{cc|cc}
        \toprule
         \textbf{BEV  Object Queries} & \textbf{Hybrid Temporal Modeling} &  \textbf{Params} & \textbf{FPS}  \\
         \midrule
         \ding{55} & \ding{55} & 69.0 M & 1.637  \\
         \ding{51} & \ding{55} & 69.0 M \green{(-0.04\%)} & 1.334 \red{(-18.5\%)}  \\
         \ding{51} & \ding{51} & 69.4 M \red{(+0.55\%)} & 1.327 \red{(-18.9\%)}  \\  
         \bottomrule
    \end{tabular}
    }
    \caption{Impact of our components on the runtime efficiency and parameter count. FPS are measured on a single NVIDIA 2080Ti using a batch size of 1 of NuScenes data.}
    \label{table:runtime}
\end{table}

\section{Conclusion}
In this paper, we demonstrated the effectiveness of \textit{DenseBEV}, a two-stage anchor generation for 3D multi-camera object detection, by interpreting each cell in a \ac{BEV} grid as a potential object. 
We also showed that fusing previous \ac{BEV} features with previously detected objects as object queries significantly enhances detection performance. 
Both \textit{DenseBEV} and \textit{DenseBEV++} were presented as lightweight extensions that substantially improved performance on nuScenes and Waymo datasets compared to the baseline.
We highlighted the importance of integrating \ac{NMS} for dense anchors, especially for detecting smaller objects.
In particular, our method excels at detecting small objects, including pedestrians and traffic cones, which are often challenging for existing approaches.
However, \ac{NMS} remains a computational bottleneck, and future work could focus on making its implementation more efficient. 
Given the small number of kept anchor boxes compared to the total number of grid cells, a customized implementation offers significant room for improvement. 
Building on our per-cell object formulation, future work could explore optimized anchor layouts, such as a polar grid, to better emphasize nearby objects and further improve detection.

\section*{Acknowledgment}
This work is a result of the joint research project STADT:up (19A22006O). The project is supported by the German Federal Ministry for Economic Affairs and Energy (BMWE), based on a decision of the German Bundestag. The author is solely responsible for the content of this publication.

{
    \small
    \bibliographystyle{ieeenat_fullname}
    \bibliography{main}
}
\clearpage
\setcounter{table}{0}
\setcounter{figure}{0}
\renewcommand{\thetable}{A\arabic{table}}
\renewcommand{\thefigure}{A\arabic{figure}}
\appendix
\newpage
   \twocolumn[
    \centering
    \Large
    \textbf{DenseBEV: Transforming BEV Grid Cells into 3D Objects \\
Supplementary Material}\\
    \vspace{1.5em}
   ] %
 
\section{Supplementary}
\subsection{Class awareness in NMS} \label{subsection:class_aware}
As stated in the paper, we refrain from using class information when applying \ac{NMS}, as early misclassifications can adversely affect performance. 
We investigate this by evaluating DenseBEV with $\tau=0.2$ on the nuScenes validation set, as shown in Table \ref{table:classaware}. 
We tested whether class-aware NMS or Scale NMS \cite{huang2021bevdet} could enhance the detection task.
Scale NMS resizes objects before calculating the \ac{IoU}, making it less intrusive than class-aware \ac{NMS}. 
It aims to better handle smaller objects, which our method already excels at. 
The results indicate that using Scale \ac{NMS} results in no significant performance difference compared to the baseline (no class-aware \ac{NMS} \& no Scale \ac{NMS}). 
Class-aware \ac{NMS}, however, leads to a substantial performance drop, with a decrease of $4.2\%$ in NDS and $5.6\%$ in mAP.
Pairing class-aware \ac{NMS} with Scale \ac{NMS} does not significantly alter the results.
Hence, we decided to use regular \ac{NMS} without any modifications.

\begin{table}[h]
    \begin{center}
        \begin{tabular}{cc|cc}
            \toprule
            Class-aware NMS & Scale NMS & \textbf{NDS} $\uparrow$ & \textbf{mAP} $\uparrow$ \\
            \midrule
            \ding{55} & \ding{55} & 0.535 & 0.433 \\
            \ding{55} & \ding{51} & 0.535 & 0.430 \\
            \ding{51} & \ding{55} & 0.493 & 0.377\\
            \ding{51} & \ding{51} & 0.495 & 0.375 \\
            \bottomrule
        \end{tabular}
    \end{center}    
    \caption{Quantitative analysis of DenseBEV on nuScenes val using different NMS methods.}
    \label{table:classaware}
\end{table}

\subsection{Performance on Large Object Detections}\label{subsection:large_obj}
We observe that using solely dense queries leads to a slight performance drop for larger objects. 
To better understand this behavior, we investigate the training dynamics of the model for these classes.
For completeness, Table 3 of the main paper reports final performance across all classes.

In Table~\ref{tab:class_perf} we show a selection of large object classes and the respective model performance at different epochs. 
The results indicate that dense queries accelerate convergence for large objects such as trucks and cars, particularly in the early epochs, where DenseBEV achieves a 4.4\% improvement for cars and a 3.0\% improvement for trucks in mAP. 
As training progresses, however, this difference diminishes, and BEVFormer ultimately achieves better results.

These findings indicate that, despite NMS, multiple dense cells may continue to compete for responsibility when handling large objects.
\begin{table}
    \centering
    \begin{tabular}{l|c|cc}
    \toprule
    Class                  & Epoch & BEVFormer mAP $\uparrow$ & DenseBEV mAP $\uparrow$ \\
                            \midrule
    \multirow{3}{*}{Truck} & 6     & 30.1 &   33.1  \\
                           & 16    & 34.1    & 35.9    \\
                           & 24    & 37.0    &  36.0   \\
                           \midrule
    \multirow{3}{*}{Car}   & 6     & 55.0     &  59.4   \\
                           & 16    & 58.8    &   61.7  \\
                           & 24    & 61.8    &  61.3   \\
                           \bottomrule
    \end{tabular}
    \caption{Class-specific performance of BEVFormer and DenseBEV (both base models) across training epochs for a selection of large object classes. The total number of training epochs was 24.}
    \label{tab:class_perf}
\end{table}

\subsection{Extended Efficiency Analysis}\label{subsection:ext_efficiency}
As computation may be limited in applications such as edge devices, we provide a more detailed overview of the runtime of our approach and potential strategies to mitigate overhead. 
The different options are summarized in Table~\ref{tab:computation}.

In runtime-critical scenarios where peak performance is not strictly required, the most effective alternative is the use of DenseBEV++-small.
The model outperforms BEVFormer-Base by more than one point in both NDS and mAP, while achieving around $50\%$ higher inference speed.

Since the overhead primarily stems from the large number of candidates in the first NMS stage before the decoder, reducing the set at this stage is an effective strategy.
We evaluated two approaches: (i) top-$k$ candidate selection and (ii) applying a confidence-based threshold. 
As shown in Table~\ref{tab:computation}, both methods significantly reduce runtime overhead while mostly maintaining performance. 
Importantly, this filtering is applied only during inference, as the model at this stage has already learned to differentiate background from objects.
While we prefer confidence-based thresholding due to its adaptive nature, both parameters can be manually tuned after training depending on the target application. 
We note, however, that the thresholding experiments were conducted solely on nuScenes, and care should be taken when applying them to different datasets or deployment settings.

\begin{table}[t]
    \centering
    \resizebox{\linewidth}{!}{
    \begin{tabular}{l|c|cc}
        \toprule
         Method & \textbf{FPS} $\uparrow$ & \textbf{NDS} $\uparrow$ & \textbf{mAP} $\uparrow$ \\
        \midrule
         BEVFormer-base & 1.637 & 51.7 & 41.6 \\
        DenseBEV++-base & 1.327 & 54.9 & 44.9 \\
        \midrule
         DenseBEV++-small & 2.449 & 52.8 & 42.8 \\
         DenseBEV++-base (top-10k) & 1.519 & 54.9 & 44.9 \\
         DenseBEV++-base (conf. 0.5\%) & 1.511 & 54.9 & 44.9 \\
         DenseBEV++-base (conf. 1\%) & 1.545 & 54.7 & 44.7 \\
         \bottomrule
    \end{tabular}
    }
    \caption{Computation comparison measured on a single NVIDIA 2080Ti using a batch size of 1 of NuScenes data. The metrics are evaluated on NuScenes val.}
    \label{tab:computation}
\end{table}

\subsection{Additional Visualizations} \label{subsection:qualitative}
\begin{figure*}[t]
    \centering
    \begin{subfigure}{\linewidth}
        \centering
        \includegraphics[width=\linewidth]{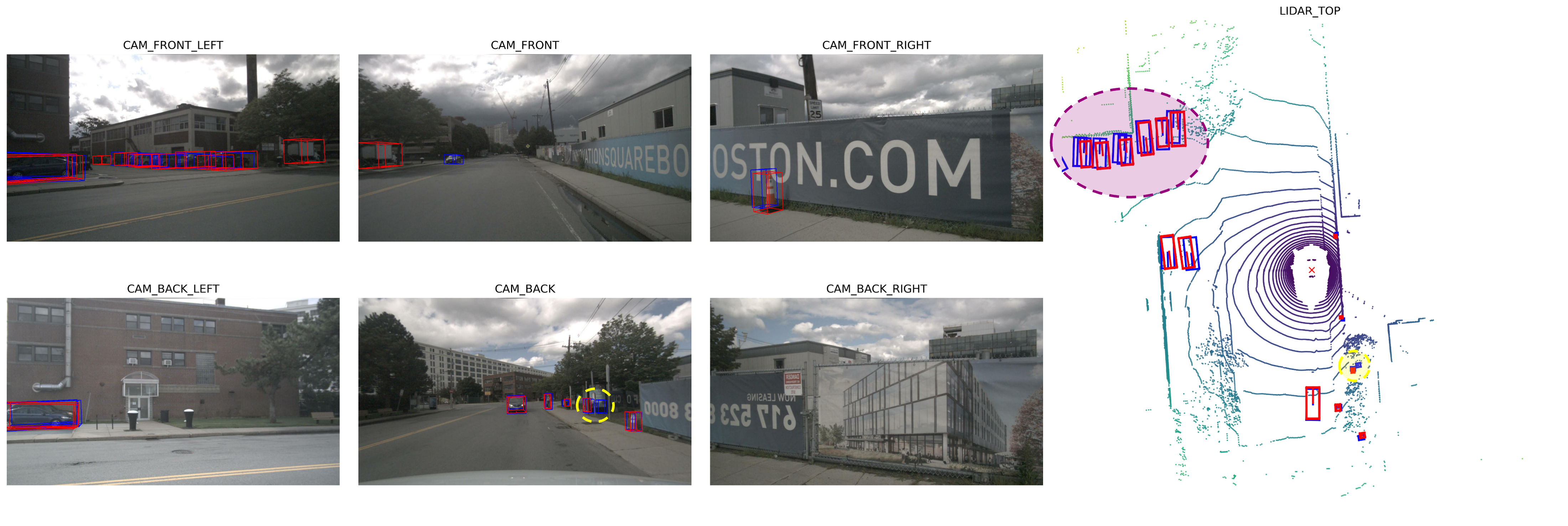}
        \caption{Visualization of baseline (BEVFormer) results.}
        \label{fig:baseline_sub}
    \end{subfigure}

    \vspace{1em} %

    \begin{subfigure}{\linewidth}
        \centering
        \includegraphics[width=\linewidth]{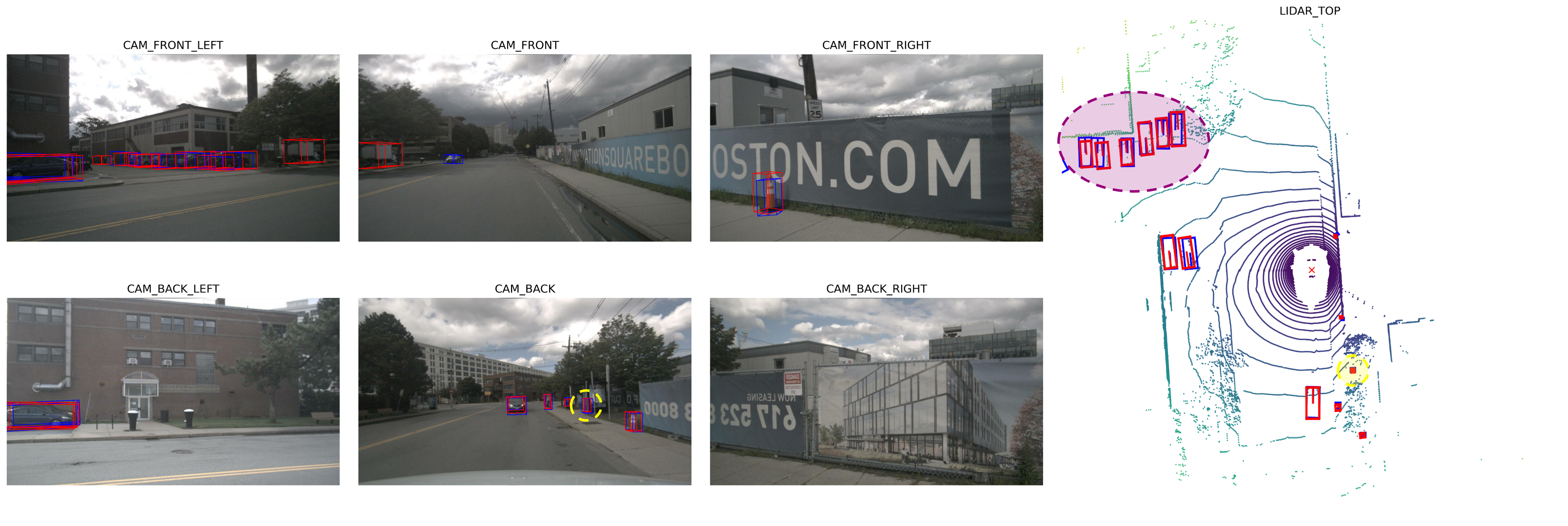}
        \caption{Visualization of DenseBEV++ results. }
        \label{fig:densebev_sub}
    \end{subfigure}
    \caption{Comparison of detection results: (a) baseline and (b) DenseBEV++. The red boxes mark the ground truth, and blue boxes the predictions. The circles highlight interesting areas in the scene.}
    \label{fig:combined_results}
\end{figure*}

In this section, we provide additional visualizations to highlight the impact of our method.
Figure \ref{fig:combined_results} shows a qualitative comparison between the baseline BEVFormer \cite{Zhiqi2022bev} and our proposed DenseBEV++.
The yellow circle in Figure \ref{fig:baseline_sub} highlights duplicate detections made by the baseline, which are effectively suppressed by our integrated \ac{NMS}, as shown in Figure \ref{fig:densebev_sub}.
The purple-highlighted area showcases how our dense anchor strategy improves localization, as demonstrated by the closer alignment between predicted and ground-truth boxes in the \ac{BEV}.

\begin{figure*}[t]
    \centering
    \begin{subfigure}{\linewidth}
        \centering
        \includegraphics[width=\linewidth]{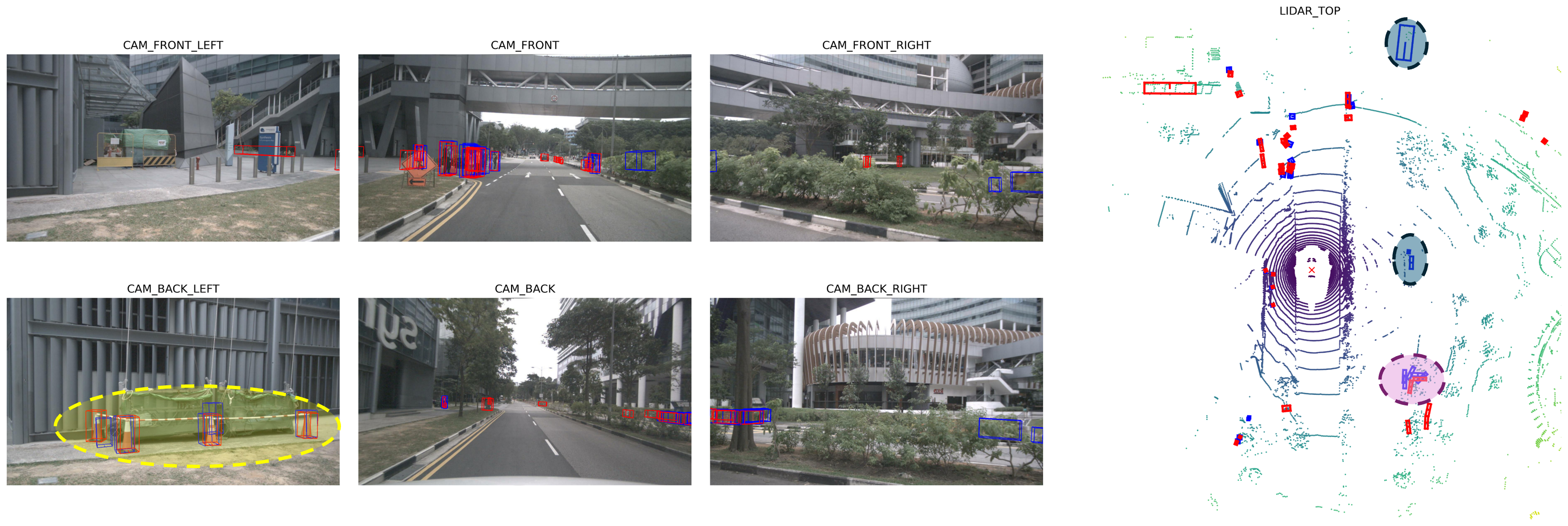}
        \caption{Visualization of baseline (BEVFormer) results.}
        \label{fig:baseline_sub1}
    \end{subfigure}

    \vspace{1em} %

    \begin{subfigure}{\linewidth}
        \centering
        \includegraphics[width=\linewidth]{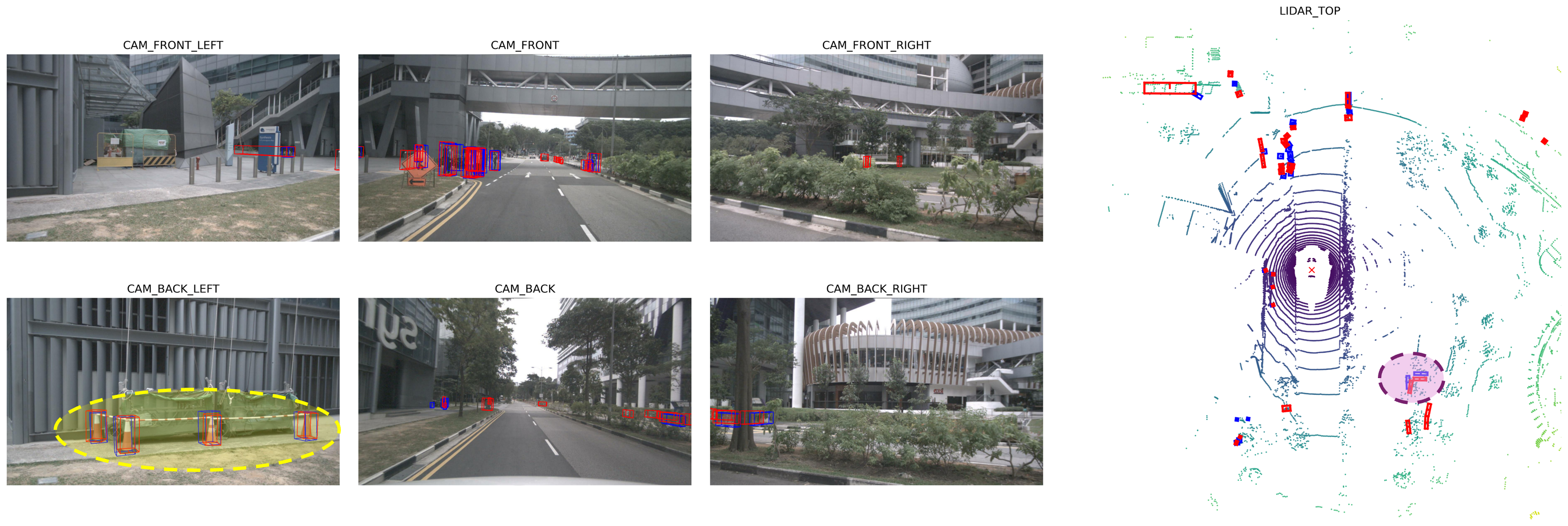}
        \caption{Visualization of DenseBEV++ results. }
        \label{fig:densebev_sub1}
    \end{subfigure}
    \caption{Comparison of detection results: (a) baseline and (b) DenseBEV++. The red boxes mark the ground truth, and blue boxes the predictions. The circles highlight interesting areas in the scene.}
    \label{fig:combined_results1}
\end{figure*}

\begin{figure*}[t]
    \centering
    \begin{subfigure}{\linewidth}
        \centering
        \includegraphics[width=\linewidth]{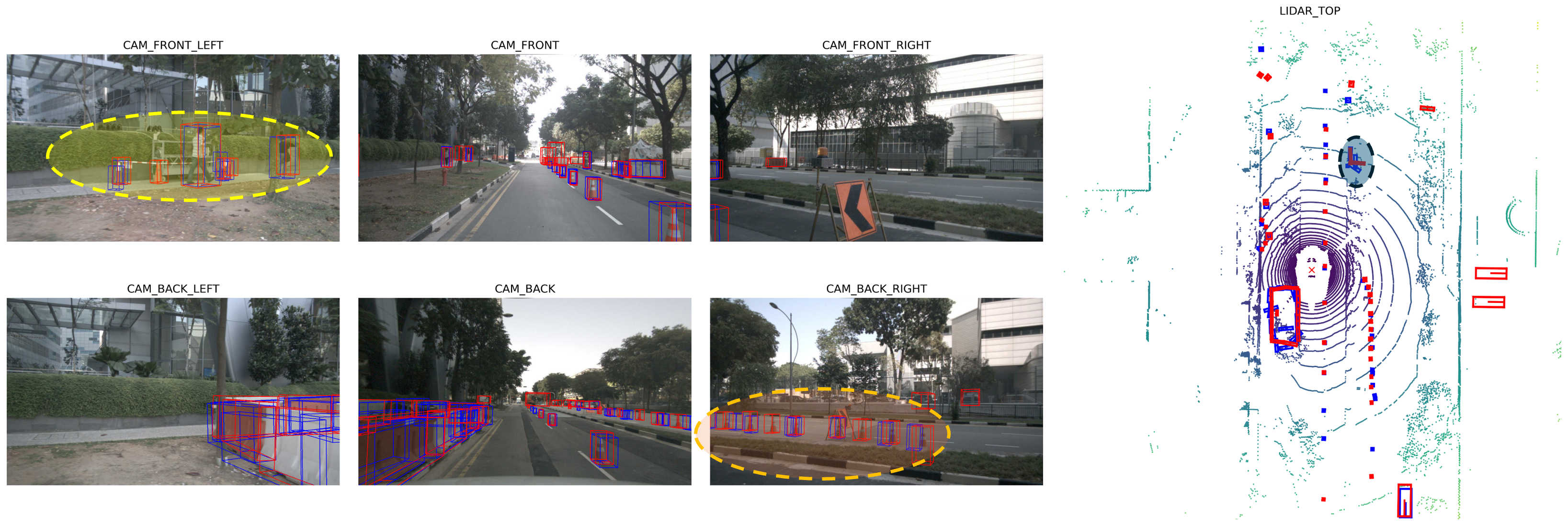}
        \caption{Visualization of baseline (BEVFormer) results.}
        \label{fig:baseline_sub2}
    \end{subfigure}

    \vspace{1em} %

    \begin{subfigure}{\linewidth}
        \centering
        \includegraphics[width=\linewidth]{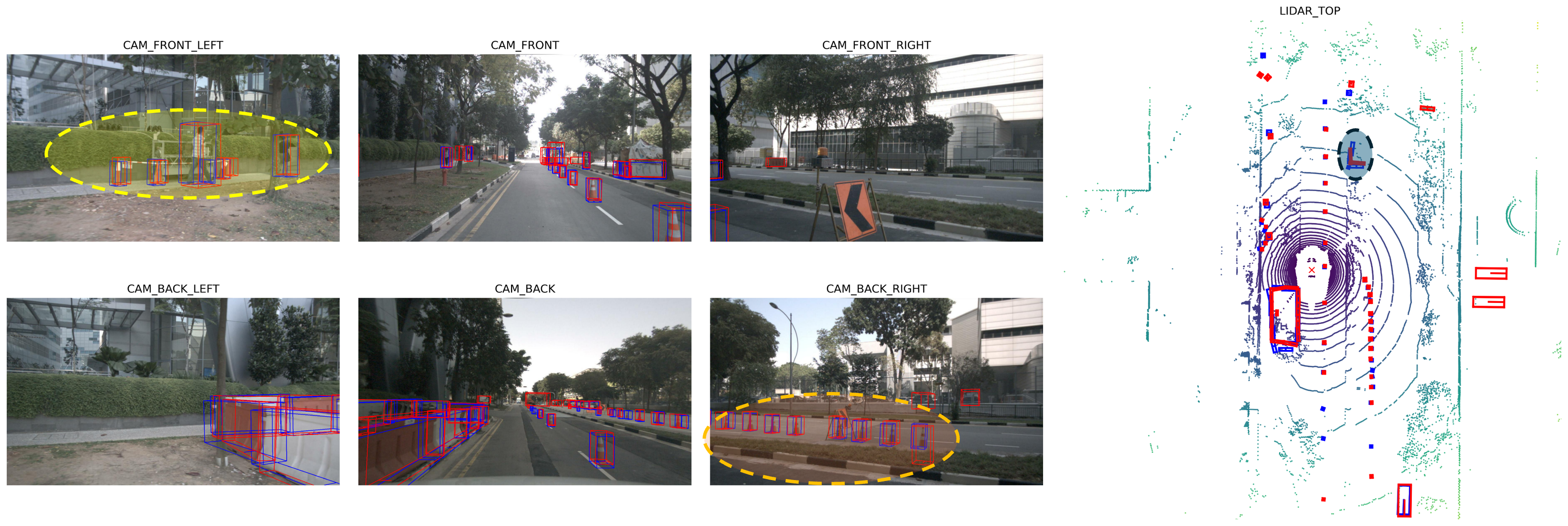}
        \caption{Visualization of DenseBEV++ results. }
        \label{fig:densebev_sub2}
    \end{subfigure}
    \caption{Comparison of detection results: (a) baseline and (b) DenseBEV++. The red boxes mark the ground truth, and blue boxes the predictions. The circles highlight interesting areas in the scene.}
    \label{fig:combined_results2}
\end{figure*}

We provide two additional examples in Figures~\ref{fig:combined_results1} and \ref{fig:combined_results2}, without discussing them in detail. The key differences are highlighted with circles. Overall, the baseline produces more false positives, especially duplicates in regions with many small objects. In contrast, \textit{DenseBEV++} yields fewer false positives, is more reliable, and detects a greater number of small objects.

\end{document}